\documentclass[conference,a4paper]{IEEEtran}
\IEEEoverridecommandlockouts
\usepackage{cite}
\usepackage{amsmath,amssymb,amsfonts,bm}
\usepackage{algorithmic}
\usepackage{graphicx}
\usepackage{textcomp}
\usepackage{xcolor}
\usepackage{algorithm}
\usepackage{tabularx,booktabs}
\usepackage{filecontents}
\usepackage{lipsum}
\usepackage{amssymb}
\usepackage{multirow}
\usepackage{adjustbox}
\usepackage{setspace}
\begin{document}

\title{Few-shot calibration of low-cost air pollution (PM$_{2.5}$) sensors using meta-learning \\

\thanks{This work has been supported by Maharashtra Pollution Control Board and Bloomberg Philanthropies}
}

\author{\IEEEauthorblockN{Kalpit Yadav$^\dagger$, Vipul Arora$^\dagger$, Sonu Kumar Jha$^\ddagger$, Mohit Kumar$^\ddagger$, Sachchida Nand Tripathi$^\ddagger$}
\IEEEauthorblockA{$^\dagger$Department of Electrical Engineering, IIT Kanpur\\
$^\ddagger$Department of NAF, IIT Kanpur\\
\{kalpit, vipular, ksonu, mohitk, snt\}@iitk.ac.in}
}

\maketitle

\begin{abstract}
Low-cost particulate matter sensors are transforming air quality monitoring because they have lower costs and greater mobility as compared to reference monitors. Calibration of these low-cost sensors requires training data from co-deployed reference monitors. Machine Learning based calibration gives better performance than conventional techniques, but requires a large amount of training data from the sensor, to be calibrated, co-deployed with a reference monitor. In this work, we propose novel transfer learning methods for quick calibration of sensors with minimal co-deployment with reference monitors. Transfer learning utilizes a large amount of data from other sensors along with a limited amount of data from the target sensor. Our extensive experimentation finds the proposed Model-Agnostic-Meta-Learning (MAML) based transfer learning method to be the most effective over other competitive baselines.
\end{abstract}

\begin{IEEEkeywords}
Air quality, low-cost sensor calibration, few-shot learning, MAML, deep learning
\end{IEEEkeywords}
\section{Introduction}
Elevated levels of air pollutants is regarded as the biggest environmental health risk in the whole world. 
Since traditional air quality monitoring stations have high setup cost and are cumbersome to maintain, administration typically only deploys a few at important locations of a city. As levels of air pollutants depends upon lot of factors and can change significantly from one location to another that are only few kilometers apart, there is an urgent need for higher resolution of data in order to properly assess people’s exposure to air pollution.

Fortunately, low-cost sensors offer an affordable solution due to their lower maintenance costs, small size and greater mobility \cite{shin2017, mitrovics2016, dutta2016, esfahani2020}. These sensors are known to be less accurate than the reference grade monitors, and hence, in-field calibration is required to obtain a reasonable sensor readings.
Two settings are commonly used, (i) calibrating the sensors against a co-deployed reference monitor, and (ii) collating the spatio-temporal measurements in a network of sensors \cite{hofman2020, you2020}. In this paper, we focus on the first setting.

Calibration involves capturing the complex non-linear relationship between the raw sensor values and the true values as measured by the reference monitors.
Various statistical methods such as ARIMA \cite{munir2021, yan2020} and Kalman filtering, and supervised machine learning methods such as linear regression, nearest neighbors and support vector regression have been studied to calibrate low cost sensor readings \cite{Sahu2020, zheng2019, holstius2014}. 
Deep learning methods such as fully connected neural network \cite{spinelle2015, spinelle2017, bagkis2021}, convolutional neural network (CNN) \cite{yu2020, bagkis2021}, recurrent neural network (RNN) \cite{han2020, bagkis2021} have also been very effective in achieving the state-of-the-art performance in sensor calibration.

In general, deep learning based methods are known to have superior performance compared to conventional machine learning methods in various learning problems \cite{goodfellow2016}. But the former typically require much larger amount of training data compared to the latter \cite{Li2006}.
One way to alleviate this problem of data scarcity is transfer learning, which leverages knowledge from abundant data from similar tasks (called as source) and fine-tunes the model over the target task \cite{naik1992, arora2016}.
A state of the art transfer learning technique is model-agnostic meta learning (MAML) \cite{finn2017, sun2018}. MAML learns an optimal initialization values for the model parameters, so that the model can quickly adapt to the target data.
Mostly these advanced deep learning methods address classification problems, while regression brings in its own challenges \cite{weinberger2007, stephane2019}.

In order to address the problem of data scarcity in sensor calibration, various transfer learning and domain adaptation methods have been proposed that attempt to transfer knowledge from a data rich source location to a target location with insufficient data. However, only few studies have been done to transfer the generalised knowledge from multiple source locations to the target location to reduce the risk of negative transfer and achieve meaningful results in the context of sensor calibration. 


This paper deals with deep learning based calibration techniques in resource-constrained setting.
We wish to minimize the amount of data from a co-located reference monitor needed to prepare the calibration model for a given sensor.
We propose a novel MAML based transfer learning method for the regression problem at hand.





\section{Our Method}
We have multiple low-cost sensors deployed at multiple locations, with one sensor per site.
For a sensor, deployed at certain location $p$, $D_p=(\bm{x}_p, \bm{y}_p)$ denote the time series of readings $\bm{x}_p$ obtained from the low-cost sensor and the corresponding readings $\bm{y}_p$ obtained from the co-deployed reference monitor, respectively. Both $\bm{x}_p$ and $\bm{y}_p$ have same length, denoted by $|D_p|$.
We have a set of source locations $P^s$, and some target locations $P^t$. Each target location has a small amount of data, i.e. $|D_{p_i}| >> |D_{p_j}| \forall p_i\in P^s, \forall p_j\in P^t$. 

We consider a model, denoted by $f_{\theta}$ with parameters $\theta$ that maps observations $x$ to output $\hat{y}$. We use a neural network to implement $f_{\theta}$.


\subsection{MAML based Sensor Calibration}

During meta training, we sample support set and query set from data $D^s_p$ of a source location $p$. The \emph{support set} is used for learning the parameters of the network, and the \emph{query set} is used for evaluating the performance of the network for that particular location. 

\subsection{Proposed Model}
Our meta learning framework learns a meta learner from multiple source locations, and uses it to train a base model at the target location, using the limited amount of training data from the target location. The base model is location specific, while the meta learner is generalized for all locations.

We implement a base model as a fully connected deep neural network represented by a function $f_{\theta_p}$ with parameters $\theta_p$. 
To effectively and quickly train the base model for any location $p$, we take help of a meta learner, which is again implemented as a full connected deep neural network, $f_{\phi}$ with parameters $\phi$. The architecture of $f_\phi$ is same as that of $f_{\theta_p}$. In MAML, we use the meta learner to initialize the base model for a new location.

First, we use the source data to train our meta learner. Then, use the limited training data from the target location along with the meta learner to train our base model at a particular target location. The meta model $f_\phi$ is used to initialize the base learner $f_{\theta_p}$ at the target location $p$.

\subsubsection{Training the Meta Learner at Source}
The meta learner $f_\phi$ is initialized randomly.
We choose a particular source location $p_i$, the data $D_{p_i}$ is split into a \emph{support set} $D^{\mathcal{S}_{p_i}}$ and a \emph{query set} $D^{\mathcal{Q}}_{p_i}$, which are both mutually exclusive. We use the support set to train the base model, while the query set is used to evaluate the base model and to train the meta learner.

The base model is initialised as $\phi$ and trained using gradient descent over $D_{p_i}^\mathcal{S}$. 
\begin{equation} \label{eq:trainBase}
\theta_{p_i} \leftarrow \theta_{p_i} - \alpha\nabla_{\theta_{p_i}} \mathcal{L}_{D_{p_i}^\mathcal{S}}(\theta_{p_i})
\end{equation}
Here, $\alpha \in \mathbb{R}$ is the learning rate.
The loss function we use is the mean absolute error, which is defined as
\begin{equation} 
\mathcal{L}_D(\theta) = \sum_{(\bm{x},\bm{y})\in D} |f_\theta(\bm{x}) - \bm{y}|
\end{equation}

The meta learner is trained on the query sets $\{D_{p_i}^\mathcal{Q}\}$ over a batch of locations $i$ sampled from source locations. The goal of this training is to optimize $\phi$ as
\begin{equation} \label{eq:trainMeta}
\phi \leftarrow \phi - \beta \sum_{i} \nabla_\phi \mathcal{L}_{D_{p_i}^\mathcal{Q}}(\theta_{p_i})
\end{equation}
Here, $\beta \in \mathbb{R}$ is the learning rate.

Finally, it is the meta learner only that is saved. The base models at different source locations are discarded.

\subsubsection{Training the Base Model at the Target Location}
To improve the prediction for a target location $p_j \in P^t$ with limited amount of training data, we initialise the network's parameters with $\phi$ and use Eq.\eqref{eq:trainBase} to train $\theta_{p_j}$.
The support set here acts as the training data, while the calibration performance is assessed using the query set.

The entire training algorithm of the calibration model is outlined as a pseudo code in Algorithm 1.
\begin{algorithm}
\caption{MAML based sensor calibration}
\begin{algorithmic}[1] 
\REQUIRE $P^s$: Set of source locations; $P^t$: set of target locations.
\REQUIRE $\alpha, \beta$: step size hyperparameters
\STATE randomly initialize $\phi$
\WHILE{not done \do}
\STATE Sample a batch of locations from $P^s$
\FOR{each location $p_i$}
\STATE Sample support datapoints $D_{p_i}^\mathcal{S} = \{x_p, y_p\}$ from $p_i$
\STATE Evaluate $\nabla_{\theta_{p_i}} \mathcal{L}_{D_{p_i}^\mathcal{S}}(\theta_{p_i})$
\STATE Compute adapted parameters with one or more gradient descent steps: $\theta_{p_i} \leftarrow  \theta_{p_i} - \alpha \nabla_{\theta_{p_i}} \mathcal{L}_{D_{p_i}^\mathcal{S}}(\theta_{p_i})$
\STATE Sample query data-points $D_{p_i}^\mathcal{Q} = \{x_p, y_p\}$
\STATE Evaluate $\mathcal{L}_{D_{p_i}^\mathcal{Q}}(\theta_{p_i})$
\ENDFOR
\STATE Update $\phi \leftarrow \phi - \beta \sum_{i} \nabla_\phi \mathcal{L}_{D_{p_i}^{\mathcal{Q}}}(\theta_{p_i})$
\ENDWHILE
\FOR{each target location $p_i \in P^t$}
\STATE Initialise the network with $\phi$
\STATE Compute the adapted parameters $\theta_p{_i}$ with few gradient descent using train set
\STATE Predict on test set
\ENDFOR
\end{algorithmic}
\end{algorithm}

\section{Experimental Results}

\subsection{Dataset Description}
We collected co-located data from 15 different locations around Mumbai city for our experiments. 
The low-cost sensors give four measurements, namely, PM$_{2.5}$, PM$_{10}$, temperature and humidity, while those from the reference monitor consist of only PM$_{2.5}$ values. 
The low-cost sensors use Nova fitness sensors, which work on the principal of laser scattering, to measure PM$_{2.5}$.
The data was collected from November 1, 2020 to March 5, 2021, with a sampling rate of 1 sample per hour.
For our transfer learning experiments, we use 10 locations as source and 5 locations, individually, as the target locations.   

From the dataset in each source location, we sample a window of 2 days for the support set and another window of 2 days as the query set. The two sets are mutually exclusive and the query set occurs later in time. 
For each target city, we sample a window of 3 days as training and validation set, and the next 15 days data for testing.

\begin{table*}[]
\renewcommand{\tabcolsep}{1.5mm}
\caption{Performance Metrics for different calibration methods, evaluated over different target locations (Loc.)}
\label{result_table}
\begin{tabular}{r|rrrr|rrrr|rrrr}
\toprule
          & \multicolumn{4}{c|}{MAE (standard deviation)}          & \multicolumn{4}{c|}{RMSE}  & \multicolumn{4}{c}{$R^2$}    \\ \hline
\multicolumn{1}{c|}{Loc.} & \multicolumn{1}{c}{$B_1$} & \multicolumn{1}{c}{$B_2$} & \multicolumn{1}{c}{$B_3$} & \multicolumn{1}{c|}{MAML} & \multicolumn{1}{c}{$B_1$} & \multicolumn{1}{c}{$B_2$} & \multicolumn{1}{c}{$B_3$} & \multicolumn{1}{c|}{MAML} & \multicolumn{1}{c}{$B_1$} & \multicolumn{1}{c}{$B_2$} & \multicolumn{1}{c}{$B_3$} & \multicolumn{1}{c}{MAML} \\ \hline
1         & 36.4 (29.1) & 13.7 (15.2) & 9.0 (18.2)  & 8.0 (6.5)   & 46.6 & 16.5 & 11.2 & 10.3 & -32.4 & 46.6  & 75.3  & 79.9 \\
2         & 21.8 (31.2) & 14.5 (27.9) & 12.8 (25.3) & 12.1 (17.5) & 32.0   & 24.3 & 21.7 & 20.2 & 0.0     & 41.4  & 53.0    & 55.7 \\
3         & 18.9 (17.1) & 12.8 (14.9) & 10.0 (17.2) & 8.7 (8.2)   & 24.7 & 15.8 & 14.1 & 12.0   & -56.7 & 46.9  & 56.3  & 68.9 \\
4         & 21.5 (22.6) & 17.1 (19.2) & 19.2 (18.3) & 15.9 (18.7) & 32.3 & 25.2 & 26.4 & 24.5 & -90.0   & -10.4 & -12.1 & 2.5  \\
5         & 18.8 (20.9) & 14.6 (19.4) & 12.6 (21.0) & 10.1 (8.2)  & 24.1 & 18.8 & 16.3 & 12.9 & 22.1  & 51.4  & 64.5  & 77.0  \\
\bottomrule
\end{tabular}
\end{table*}


\subsection{Implementation Details}
For the proposed method, we model $f_\theta$ as a standard feed-forward neural network. It consists of two hidden layers, each of which has 128 neurons, each with ReLU activation. The output layer has single linear neuron. For training the base model, we use stochastic gradient descent (SGD) optimizer with a learning rate of $10^{-3}$, and for training the meta learner, we use Adam optimizer with a learning rate of $10^{-4}$. 

\subsection{Baselines}
We compare our proposed MAML based model with three baselines methods.
The first baseline, $B_1$, uses the limited amount of training data from target location to train the calibration model without any transfer. This is a common training method employed in basic deep learning. The second baseline, $B_2$, trains the model using training data from a single source location and then transfers the model by fine-tuning it on the target location. This fine-tuning uses Adam optimization and stops when the validation loss stops improving. The third baseline, $B_3$, trains the model using combined data from all the source locations and then fine-tunes the model on a target location as in $B_2$.

\subsection{Performance metrics}
The performance of a calibration model is assessed by comparing the calibrated PM$_{2.5}$ values against the corresponding readings from the co-located reference monitor.
The metrics used for this comparison are mean absolute error (MAE), root mean square error (RMSE) and the coefficient of determination ($R^2$). We also compute the standard deviation corresponding to the mean absolute error.


\subsection{Results}
\begin{figure} 
	\centering
	\begin{tabular}{c}
		\includegraphics[width=\linewidth]{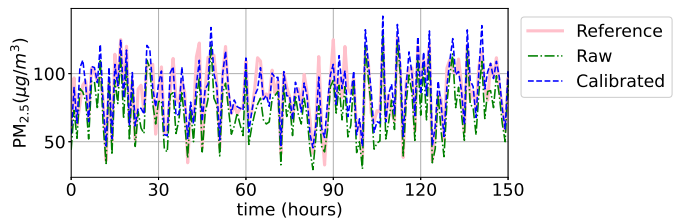}\\
	\end{tabular}
	\caption{Example of PM$_{2.5}$ readings from a low-cost sensor (raw), the co-located reference monitor and the proposed calibration model.}
	\label{fig_timeseries}
\end{figure}

We implement our method and compare the calibrated PM$_{2.5}$ values using different methods for five target cities. The performance of different calibration methods is shown in Table~\ref{result_table}. The baseline $B_1$ performs the worst, showing that deep learning is not effective when the training data is small. The baseline $B_2$ performs much better as compared to $B_1$ illustrating the benefits of transfer learning. The performance of transfer learning improves further with the increase in the amount of source data, as supported by the improved results with $B_3$. However, the proposed method outperforms all others, showing that the proposed MAML based transfer learning is better than fine-tuning based transfer learning. While the proposed method shows improvements in MAE, RMSE and R$^2$ values, we also notice a decline in the standard deviation of absolute error, indicating that the calibrated values follow the reference monitor more closely.
Fig.~\ref{fig_timeseries} illustrates an example of the calibrated PM$_{2.5}$ values against the raw values from low-cost sensors and readings from reference monitors. We can see the calibrated values follow the reference monitor more closely as compared to the raw sensor readings. Code is available on Github\footnote{\url{https://github.com/madhavlab/2021KalpitBTMT}}.



\section{Conclusion and Future Work}
This paper addresses the problem of quick and efficacious calibration of low cost sensors by minimal co-deployment with reference monitors.
It uses deep learning for efficacy and transfer learning for minimal co-deployment.
The main contributions of this paper are (i) an efficient solution for calibration of low-cost sensors, and (ii) a model agnostic meta learning based transfer learning method for regression.
The proposed method is shown to outperform other learning schemes in the transfer learning setting.
The proposed scheme is useful for calibration of static sensors with mobile reference monitors or that of mobile sensors with static reference monitors.
In future, with longer deployments, we wish to address sensor drift and maintenance issues.
The proposed method is not limited to PM$_{2.5}$ calibration only. We plan to extend our model for calibration of various gas sensors such as those for NO$_x$, O$_3$ and CO.
We also plan to extend our work to calibration of dense sensor network to monitor the air quality. 





\clearpage

\begin{thebibliography}{00}

\bibitem{you2020} Y. You, A. Xu, and T. J. Oechtering. ``Belief Function Fusion based Self-calibration for Non-dispersive Infrared Gas Sensor.'' In 2020 IEEE Sensors, pp. 1-4. IEEE, 2020.
\bibitem{shin2017} S.H. Shin, and S. Lee. "Multi-channel multi-mode ROIC with embedded calibrations for environmental gas sensors." in IEEE SENSORS, 2017.
\bibitem{dutta2016} J. Dutta, F. Gazi, S. Roy, and C. Chowdhury. ``AirSense: Opportunistic crowd-sensing based air quality monitoring system for smart city,'' In 2016 IEEE SENSORS, pp. 1-3. IEEE, 2016.
\bibitem{esfahani2020} S. Esfahani, P. Rollins, J. P. Specht, M. Cole, and J. W. Gardner. "Smart City Battery Operated IoT Based Indoor Air Quality Monitoring System." In 2020 IEEE Sensors, pp. 1-4. IEEE, 2020.
\bibitem{zheng2019} T. Zheng, M. H. Bergin, R. Sutaria, S. N. Tripathi, R. Caldow, and D. E. Carlson, ``Gaussian process regression model for dynamically calibrating and surveilling a wireless low-cost particulate matter sensor network in Delhi,'' Atmos. Meas. Tech., vol. 12, no. 9, pp. 5161–5181, 2019.
\bibitem{Sahu2020} R. Sahu, et al., ``Robust statistical calibration and characterization of portable low-cost air quality monitoring sensors to quantify real-time O$_3$ and NO$_2$ concentrations in diverse environments,'' Atmos. Meas. Tech. Discuss., no. June, 2020.
\bibitem{hofman2020} J. Hofman, M. E. Nikolaou, T. H. Do, X. Qin, E. Rodrigo, W. Philips, N. Deligiannis, V.P. La Manna, ``Mapping air quality in IoT cities: cloud calibration and air quality inference of sensor data,'' IEEE Sensors, 2020
\bibitem{mitrovics2016} J. Mitrovics, "Smart sensors for air quality monitoring: Concepts and new developments." In 2016 IEEE SENSORS, pp. 1-2. IEEE, 2016.
\bibitem{finn2017} C. Finn, P. Abbeel, and S. Levine. ``Model-agnostic meta-learning for fast adaptation of deep networks,'' in ICML, 2017.
\bibitem{sun2018} Q. Sun, Y. Liu, T. S. Chua, and B. Schiele, ``Meta-transfer learning for few-shot learning,'' in IEEE CVPR, 2018, pp. 403–412.
\bibitem{naik1992} D.K. Naik and R. J. Mammone. ``Meta-neural networks that learn by learning.'' IEEE International Joint Conference on Neural Networks, 1992.
\bibitem{arora2016} V. Arora, A. Lahiri, and H. Reetz, ``Attribute based shared hidden layers for cross-language knowledge transfer,'' in IEEE Spoken Language Technology Workshop (SLT), 2016.
\bibitem{goodfellow2016} Goodfellow, Ian, Yoshua Bengio, and Aaron Courville. ``Deep learning,'' MIT press, 2016.
\bibitem{Li2006} F. Li, R. Fergus, and P. Perona. ``One-shot learning of object categories,'' IEEE Trans. Pattern Anal. Mach. Intell., 28(4):594–611, 2006.
\bibitem{weinberger2007} K. Q. Weinberger and G. Tesauro, “Metric learning for kernel regression,” J. Mach. Learn. Res., vol. 2, pp. 612–619, 2007.
\bibitem{stephane2019} S. Lathuilière, P. Mesejo, X. Alameda-Pineda, and R. Horaud. "A comprehensive analysis of deep regression." IEEE transactions on pattern analysis and machine intelligence, vol. 42, no. 9: 2065-2081, 2019.
\bibitem{holstius2014}  D. Holstius, A. Pillarisetti, K. R. Smith, and E. Seto. “Field calibrations of a low-cost aerosol sensor at a regulatory monitoring site in California.” Atmospheric Measurement Techniques, vol. 7: 1121-1131, 2014.
\bibitem{han2020} Han, P.; Mei, H.; Liu, D.; Zeng, N.; Tang, X.; Wang, Y.; Pan, Y. Calibrations of Low-Cost Air Pollution Monitoring Sensors for CO, NO2, O3, and SO2. Sensors 2021, 21, 256.
\bibitem{spinelle2015} L. Spinelle, M. Gerboles, M. G. Villani, M. Aleixandre, and F. Bonavitacola, ``Field calibration of a cluster of low-cost available sensors for air quality monitoring. part a: Ozone and nitrogen dioxide,'' Sensors and Actuators B: Chemical 215:249–257, 2015.
\bibitem{spinelle2017} L. Spinelle, M. Gerboles, M. G. Villani, M. Aleixandre, and F. Bonavitacola, ``Field calibration of a cluster of low-cost commercially available sensors for air quality monitoring. part b: NO, CO and CO2,'' Sensors and Actuators B: Chemical 238:706–715, 2017.
\bibitem{yu2020} H. Yu, Q, Li, Y.-A. Geng, Y. Zhang, Z. Wei, ``AirNet: A Calibration Model for Low-Cost Air Monitoring Sensors Using Dual Sequence Encoder Networks,'' in AAAI, 2020
\bibitem{munir2021} S. Munir, M. Mayfield, ``Application of Density Plots and Time Series Modelling to the Analysis of Nitrogen Dioxides Measured by Low-Cost and Reference Sensors in Urban Areas,'' Nitrogen, 2, 167–195, 2021.
\bibitem{yan2020} X. Yan and L. Shuangting ``Time Series and Multiple Linear Regression Calibration Model for CO Monitoring Data,'' in E3S Web of Conferences, vol. 214, p. 03039. EDP Sciences, 2020.
\bibitem{bagkis2021} E. Bagkis, T. Kassandros, M. Karteris, A. Karteris, K. Karatzas ``Analyzing and Improving the Performance of a Particulate Matter Low Cost Air Quality Monitoring Device". Atmosphere 2021, 12, 251.

\end{thebibliography}
\end{document}